# Always-On, Sub-300-nW, Event-Driven Spiking Neural Network based on Spike-Driven Clock-Generation and Clock- and Power-Gating for an Ultra-Low-Power Intelligent Device

Dewei Wang[1], Pavan Kumar Chundi[1], Sung Justin Kim[1], Minhao Yang[1], Joao Pedro Cerqueira[1],
Joonsung Kang[2], Seungchul Jung[2], Sangjoon Kim[2], and Mingoo Seok[1]
Columbia University[1], Samsung Electronics[2]
E-mail: dewei.wang@columbia.edu

*Abstract*—Always-on artificial intelligent (AI) functions such as keyword spotting (KWS) and visual wake-up tend to dominate total power consumption in ultra-low power devices [1]. A key observation is that the signals to an always-on function are sparse in time, which a spiking neural network (SNN) classifier can leverage for power savings, because the switching activity and power consumption of SNNs tend to scale with spike rate. Toward this goal, we present a novel SNN classifier architecture for always-on functions, demonstrating sub-300nW power consumption at the competitive inference accuracy for a KWS and other always-on classification workloads.

## I. INTRODUCTION

SNN classifiers have been attracting a large amount of attention for ultra-low-power intelligence. Especially, the asynchronous versions are promising for always-on functions thanks to event-driven operation, which enables power dissipation proportional to input rate. However, most of the existing asynchronous SNNs [2] employ complex logic such as quasi-delay-insensitive (QDI) dual-rail dynamic logic, which is significantly bulkier and power-hungrier than single-rail static counterpart and also not very voltage-scalable [3,4]. On the other hand, synchronous SNNs use power-efficient static logic [5], but they target high throughput, not always-on function, and focus on minimizing *energy* consumption, not power. As a result, they exhibit the power consumption of tens of mW.

In this work, therefore, we propose an always-on SNN classifier consuming <300 nW. Our architecture uses only static logic operating at near-threshold voltage (NTV) while being fully spike-event-driven. Specifically, we design *the neurosynaptic core in static gates and equip it with i) spatiotemporally fine-grained clock-generation, ii) clock-gating and iii) power-gating, all driven by spikes*. Also, the communication fabric between neurosynaptic cores is simply wires and yet free of spike collision. This event-driven architecture incurs zero switching at no input change, thereby exhibits power consumption proportional to input rate.

We prototyped a 5-layer SNN classifier having 650 neurons and 67,000 synapses in a 65-nm CMOS. The SNN hardware demonstrates 7 to 1000X less power consumption at the state-of-the-art accuracies for well-known KWS benchmarks: Google Speech Command Dataset (GSCD) for multi-keyword recognition [6] and HeySnips for single-keyword spotting [7].

## II. SPIKE-EVENT-DRIVEN SNN ARCHITECTURE

Fig. 1 shows the proposed SNN classifier. It has five neurosynaptic cores and maps a fully-connected SNN model as large as 256-128-128-128-10. Each core contains a neuron block and a synapse block (the last layer has no synapse). A neuron block contains up to 256 integrate-and-fire (IF) neurons. A synapse block has i) an arbiter, ii) an SRAM storing up to 256-by-128 binary weights, and iii) a spike generator that simultaneously generates 128 spikes. The communication fabric that spikes travel is simple wires. The arbiter in a synapse block ensures no spike collision in the communication fabric.

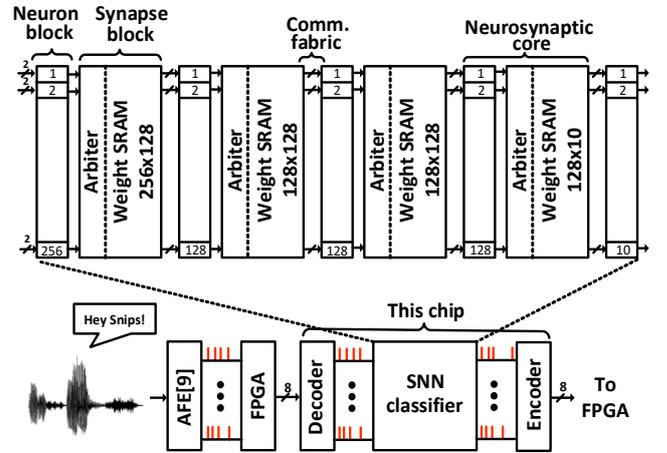

**Fig. 1. Proposed SNN architecture and testing setup**

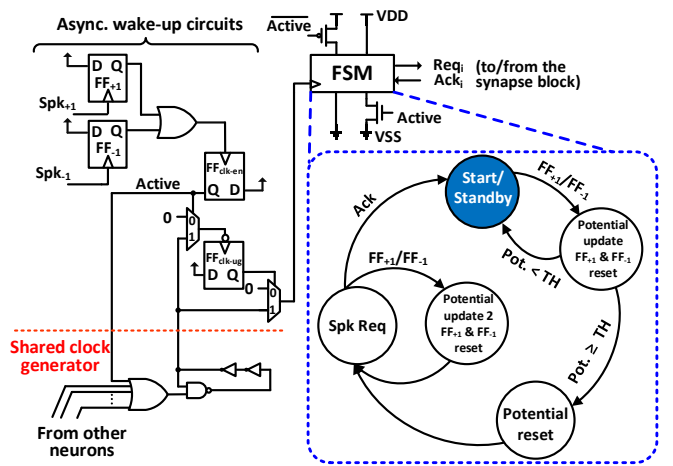

**Fig. 2. Proposed neuron block architecture**

We devise a fully spike-event-driven architecture for IF neurons (Fig. 2). It employs fine-grained clock-generation and clock-gating circuits. Also, the non-retentive parts of neurons employ zigzag power-gating switches (PGSs) [8] for leakage suppression. Each neuron has i) *asynchronous* wake-up circuits and ii) a *synchronous* finite state machine (FSM).

The wake-up circuits (Fig. 2 left) can detect the rising edge of incoming spike signals. There are two inputs, $spk_{+1}$ incrementing and $spk_{-1}$ decrementing neuron's potential, which are detected by using static flip-flops, $FF_{+1}$ and $FF_{-1}$. The detection of a spike sets the clock-enable flip-flop ($FF_{clk-en}$), making its output *Active* high. This starts up the shared clock generator in the neuron cluster if it has not been started up by other neurons. It also un-gates the zigzag PGS of the FSM in a single cycle. Then, the first falling edge of clock generator's output sets $FF_{clk-ug}$, un-gating the clock signal that goes to the FSM. The use of $FF_{clk-ug}$ ensures that clock starts from the low phase, giving the sufficient setup time to the flip-flops in the FSM. Indeed, all the other neurons in the cluster that receive no spike remain clock- and power-gated, saving power.

Once waken up, the neuron's FSM executes the IF neuron model (Fig. 2 right). In the *Start/Standby* state, it resets $FF_{clk-en}$ and $FF_{clk-ug}$ in the wake-up circuits, which gates clock and power. When either of $FF_{+1}$ or $FF_{-1}$ receives a new spike, the FSM enters the *Potential update* state. The neuron's potential is increased or decreased by one based on the input spike's type and $FF_{+1}$ and $FF_{-1}$ are reset for readily receiving next spike. Then, the neuron's potential is compared with the preset threshold. If the potential is less than the threshold, the FSM goes back to the *Start/Standby* state. Otherwise it resets the neuron's potential to zero and assert firing request ($Req_i$). While waiting for the acknowledgement ($Ack_i$) from the arbiter in the synapse block, we add the state, *Potential update 2*, such that the FSM can receive a new spike. Once acknowledged, the neuron's FSM goes back to the *Start/Standby* state.

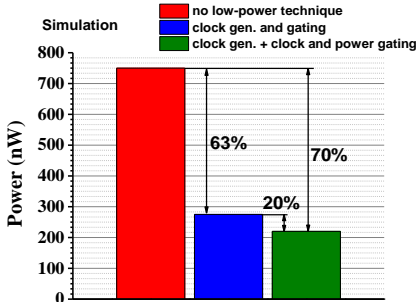

**Fig. 3.** Power savings of spike-driven clock-gen., -gating, & power-gating

This spike-driven clock generation, gating, and power gating enables large power savings. For the targeted benchmarks, the longest idle time between two spikes per neuron is estimated ~4ms at the maximum input rate. The SNN hardware without the spike-driven power management would consume 750nW (Fig.3). The proposed clock-generation/-gating enables 63% power savings and the zigzag power gating provides additional 20%, resulting in the overall power reduction of 70%. Note that the actual power savings are expected much greater since the average idle time of neurons would be orders of magnitude lower than the maximum one considered above.

We also design the synapse block to be fully spike-event-driven (Fig. 4). A request from neurons ($Req_i$) starts up its local clock generator and thus the arbiter FSM. In case more than one neuron assert $Req_i$, the arbiter FSM arbitrates the access of the single-port weight SRAM. The weight SRAM stores the 128 binary weights in the n-th row, which the n-th neuron in the current neurosynaptic core drives to the post-synaptic neurons. Therefore, upon the n-th neuron's request, the arbiter asserts the n-th wordline ($WL_n$) and loads the binary weights on the read-bitlines (RBLs). The spike generator then takes RBLs and generate 128 positive or negative spikes to the next neurosynaptic core. The arbiter then acknowledges back to the neuron by asserting $Ack_n$. If there is any outstanding $Req_i$, the arbiter FSM continue to serve, otherwise, the clock is disabled.

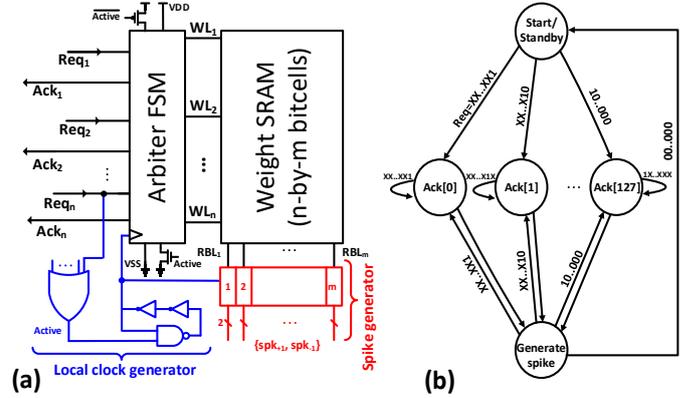

**Fig. 4.** (a) Proposed synapse block architecture and (b) the arbiter FSM

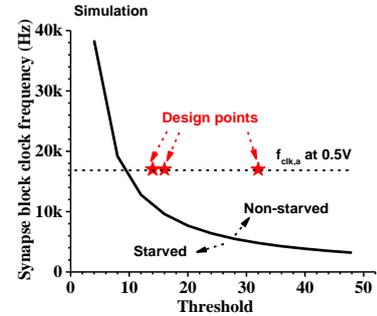

**Fig. 5.** Threshold and clock frequency optimization for no starvation

We design the arbiter with a fixed-priority scheme (Fig. 4(b)). In our experiment, it requires 23X smaller silicon area than a round-robin one. The chosen priority scheme, however, could cause the neuron with the lowest priority to starve. Therefore, we analyze and optimize several design parameters to eliminate such a starvation. We first formulate the number of requests in the *i*-th layer ($N_{req,i}$) to:

$$N_{req,i} = \frac{N_{spk,i} \times N_{nrn,i}}{TH_i}, \quad (1)$$

where $N_{spk,i}$ is the number of incoming spikes per frame (a time duration in which a feature vector is generated) and per neuron in the *i*-th neurosynaptic core, $N_{nrn,i}$ is the number of neurons in the *i*-th neurosynaptic core, and $TH_i$ is the threshold of neurons in the *i*-th neurosynaptic core. On the other hand, the number of requests that the arbiter is able to serve ($N_{serve,i}$) can be formulated to:

$$N_{serve,i} = \frac{f_{clk,a} \times T_{frame}}{N_{cyc,a}}, \quad (2)$$

where $N_{cyc,a}$ is the number of cycles that the arbiter consumes to serve one request, $T_{frame}$ is the frame length, $f_{clk,a}$ is the arbiter's clock frequency.

If $N_{req,i}$ (Eq. 1) exceeds $N_{serve,i}$ (Eq. 2), starving happens. To eliminate such, we can increase $TH_i$ and arbiter clock frequency ($f_{clk,a}$) (Fig. 5). The former, however, can reduce the number of spikes generated in the $i$-th layer and thus incur accuracy degradation. The latter can increase the power consumption of the synapse block. We thus swept $TH_i$ and $f_{clk,a}$ and found several optimal points which are used in this chip (Fig. 5).

It is noteworthy that the arbiter also enables the use of wires for spike communication. If two (post-synaptic) spikes travel to a single neuron at the same time, they collide, causing information loss. The arbiter systematically eliminates such collisions by guaranteeing that a post-synaptic neuron receives only one spike at a time.

III. EXPERIMENT AND MEASUREMENT

We prototyped the test chip in a 65nm CMOS (Fig. 6). We added the input decoder and the output encoder for reducing chip I/O counts (Fig. 1). The SNN is envisioned to interface directly with a spike-generating feature-extraction front end such as [9,10]. For the experiment, we generate 16-dimension features from [9] and feed them to the SNN using an FPGA based interface (Fig. 1). The central frequencies of the 16 channels are geometrically scaled from about 100 to 5kHz. In GSCD and HeySnips datasets, each keyword audio sample is roughly 1 s. We set the frame length $T_{frame}$ at 80ms without frame overlapping. We send the current frame together with the past 15 frames to SNN classifier (i.e., the input dimension is 256). We configure the front end [9] to generate 6-bit features for each frame.

We train the SNN equivalent binary neural network (BNN) model that uses binary weights (+1, -1) and 6-bit ReLU activation (Fig. 7) [11]. This provides the weights for the SNN model. The 6-bit activations in the BNN are spike-rate encoded for the SNN, e.g., $010000_{(2)}$ is mapped to 16 spikes/frame. In the SNN model, as spikes pass through the neurons in a layer, the number of spikes scales roughly by the ratio of the threshold. We thus set the threshold of the neurons in each layer such that each neuron generates at most 63 spikes/frame, matched to the 6-bit activation of the BNN model. Note that we can easily change the activation bit count for different models by configuring the thresholds. For example, we use 8-bit activation with the $T_{frame}$ of 0.5s for the MNIST grayscale.

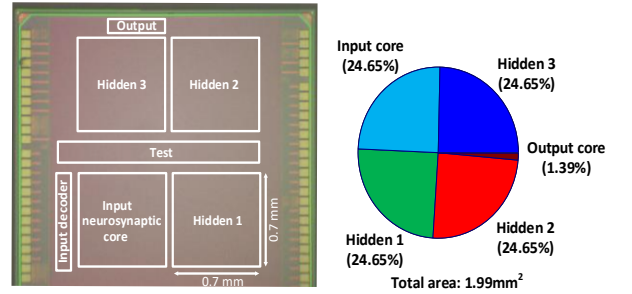

Fig. 6. Die photo and area break-down

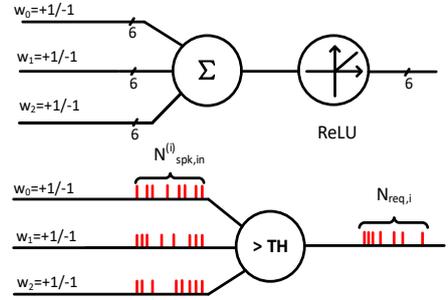

Fig. 7. Binary Coding in a BNN and spike-rate coding in an SNN

At our target voltage (0.5V), the neuron block operates at 70kHz and the synapse block operates at 17kHz (Fig. 8). The proposed SNN achieves 75-220nW power dissipation that scales with the input rate (Fig. 9). Fig. 10 shows the accuracy performance across several benchmarks. In GSCD, the SNN can recognize four keywords ("yes", "stop", "right", and "off", arbitrarily chosen) and fillers. The SNN has the architecture of 256-128-128-128-5 with the thresholds of (1,28,18,10). In HeySnips, it can recognize one keyword ("Hey Snips") and fillers. And in MNIST grayscale, we reduced the image size to 16x16 by utilizing 2x2 max-pooling. The trained SNN structure is 256-128-128-128-10 with the thresholds of (1,24,12,8).

Fig. 11 shows the receiver operating characteristic (ROC) curve for GSCD and HeySnips. False reject rate (FRR) vs. the false alarm rate (FAR) under 1-hour-long audio concatenated by test set samples are presented. In addition, we mix the speech audio with the white noise at various SNR We adopt the noise-dependent training [9] in this experiment. The SNN achieves reasonably high accuracy across 0 to 40dB SNR levels (Fig. 12). Finally, the data precision can be traded off for power savings (Fig. 13).

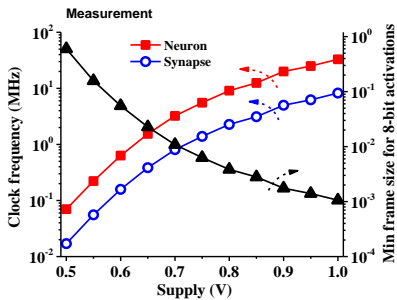

Fig. 8. Clock frequency and frame length

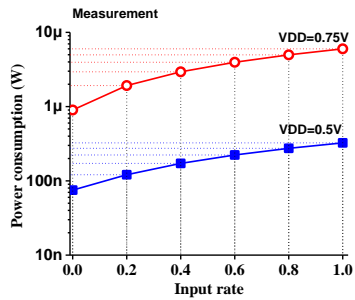

Fig.9. Power consumption vs. input rate

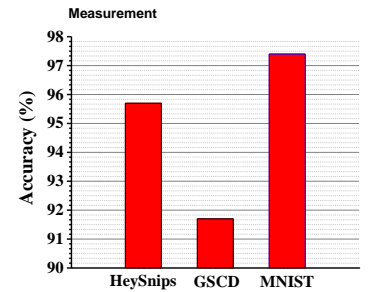

Fig.10. Accuracies across multiple benchmarks

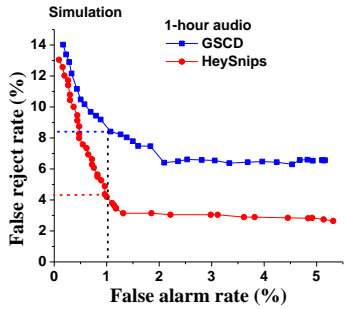
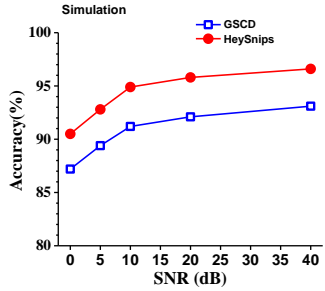
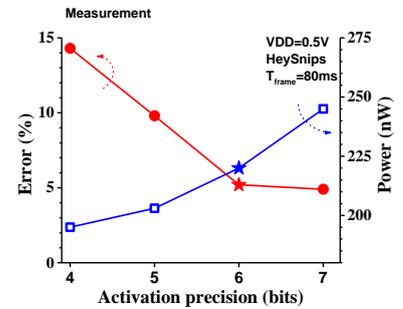

Fig.11. ROC curves from KWS benchmarks   Fig.12. Accuracy across 0 to 40dB SNR levels   Fig.13. Activation precision vs. power

We compare our work to the recent KWS accelerators (Table I) and SNN hardware (Table II). Our design is one of few SNNs targeting always-on function, achieving 7 to 1,000X power savings at the competitive accuracies in both of the benchmarks.

**Table I. Comparisons with recent KWS hardware**

|  | This work | Shan ISSCC20[12] | Guo VLSI19[13] | Giraldo ESSCIRC18[14] |
|---|---|---|---|---|
| Technology [nm] | 65 | 28 | 65 | 65 |
| Algorithm | SNN | DSCNN | RNN | LSTM |
| Area[mm$^2$] | 1.99 | 0.23 | 6.2 | 1.035 |
| VDD[V] | 0.5-1 | 0.41 | 0.9-1.1 | 0.575 |
| Clock frequency | 70KHz@0.5V | 40KHz | 75MHz | 250KHz |
| Benchmark | GSCD (4 Keywords) | GSCD (2 Keywords) | GSCD (10 Keywords) | TIMIT (4 Keywords) |
| Accuracy[%] | 91.8 | 94.6 | 90.2 | 92.0 |
| Additional benchmark | HeySnips (1 Keyword) | GSCD (1 Keyword) | HeySnips (1 Keyword) | N/A |
| Accuracy[%] | 95.8 | 98.0 | 91.9 | N/A |
| Power | 75-220nW* | 510nW** | 134µW | 5µW |

\* Power consumption scales with input rate;   \*\* Feature extraction circuits included.

**Table II. Comparisons with recent SNN hardware**

|  | This work | Park ISSCC19[5] | Chen VLSI18[15] | TrueNorth[2] |
|---|---|---|---|---|
| Technology [nm] | 65 | 65 | 10 | 28 |
| Neuron count | 650 | 410* | 4096 | 1M |
| Synapse count | 67K | N/A | 1M | 256M |
| Area[mm$^2$] | 1.99 | 10.08 | 1.72 | 430 |
| Clock frequency | 70KHz@0.5V | 20MHz | 105MHz@0.5V | N/A |
| MNIST Classification | | | | |
| Power | 305nW | 23.6mW | 9.42mW** | 63mW |
| Accuracy[%] | 97.6 | 97.8 | 97.9 | 97.6*** |
| Throughput [infs/s] | 2 | 100K | N/A | N/A |
| Energy per inference [nj] | 195 | 236 | 1,700 | N/A |
| Energy per SOP [pj] | 1.5 | N/A | 3.8 | 26 |

\* Input layer not included; \*\* Estimated from neuron's power dissipation
\*\*\* Estimated from Hsin-Pai Cheng et al, IEEE DATE 2017

## IV. CONCLUSION

In this paper, we present a fully spike-event-driven SNN classifier for always-on intelligent function. By taking advantage of signal sparseness, the SNN hardware consumes 75 to 220 nW. We train the SNN for multiple always-on functions, notably multi- and single-keyword spotting benchmarks across SNR levels, achieving competitive accuracies.

ACKNOWLEDGEMENT

This research is in part supported by Samsung Electronics and DARPA (the µBrain program).

REFERENCES

[1] K. Badami et al., "Context-aware hierarchical information-sensing in a 6µW 90nm CMOS voice activity detector." IEEE International Solid-State Circuits Conference-(ISSCC), 2015

[2] P.A. Merolla et al., "A million spiking-neuron integrated circuit with a scalable communication network and interface." Science, 2014

[3] Yu Chen, Mingoo Seok, Steve M. Nowick, "Robust and Energy-Energycient Asynchronous Dynamic Pipelines for Ultra-Low-Voltage Operations Using Adaptive Keeper Control," IEEE ACM International Symposium on Low Power Electronics and Design-(ISLPED), 2013

[4] Jian Liu, Steve M. Nowick, Mingoo Seok, "Soft MOUSETRAP: a Bundled-Data Asynchronous Pipeline Scheme Tolerant to Random Variations at Ultra Low Supply Voltages," IEEE International Symposium on Asynchronous Circuits and Systems-(ASYNC), 2013

[5] Jeongwoo Park, Juyun Lee, and Dongsuk Jeon. "A 65nm 236.5 nJ/classification neuromorphic processor with 7.5% energy overhead on-chip learning using direct spike-only feedback." IEEE International Solid-State Circuits Conference-(ISSCC), 2019.

[6] Pete Warden, "Speech Commands: A Dataset for Limited -Vocabulary Speech Recognition", arXiv: 1804.03209 [cs. CL]

[7] Alice Coucke et al., "Efficient keyword spotting using dilated convolutions and gating." IEEE International Conference on Acoustics, Speech and Signal Processing-(ICASSP), 2019

[8] J. P. Cerqueira and M. Seok, "Temporally Fine-Grained Sleep Technique for Near- and Subthreshold Parallel Architectures," in IEEE Transactions on Very Large Scale Integration (VLSI) Systems, Jan. 2017

[9] M. Yang, C. Yeh, Y. Zhou, J. P. Cerqueira, A. A. Lazar and M. Seok, "Design of an Always-On Deep Neural Network-Based 1- µW Voice Activity Detector Aided With a Customized Software Model for Analog Feature Extraction," in IEEE Journal of Solid-State Circuits-(JSSCC), June 2019

[10] M. Yang, S. Liu and T. Delbruck, "A Dynamic Vision Sensor With 1% Temporal Contrast Sensitivity and In-Pixel Asynchronous Delta Modulator for Event Encoding," in IEEE Journal of Solid-State Circuits-(JSSCC), Sept. 2015

[11] Yongqiang Cao et al., "Spiking deep convolutional neural networks for energy-efficient object recognition." International Journal of Computer Vision-(IJCV), 2015

[12] W. Shan et al., "14.1 A 510nW 0.41V Low-Memory Low-Computation Keyword-Spotting Chip Using Serial FFT-Based MFCC and Binarized Depthwise Separable Convolutional Neural Network in 28nm CMOS," IEEE International Solid- State Circuits Conference - (ISSCC), 2020

[13] R. Guo et al., "A 5.1pJ/Neuron 127.3us/Inference RNN-based Speech Recognition Processor using 16 Computing-in-Memory SRAM Macros in 65nm CMOS," IEEE Symposium on VLSI Circuits, 2019

[14] Juan SP Giraldo, and Marian Verhelst. "Laika: A 5uW programmable LSTM accelerator for always-on keyword spotting in 65nm CMOS." ESSCIRC IEEE European Solid State Circuits Conference-(ESSCIRC), 2018

[15] G. K. Chen, R. Kumar, H. E. Sumbul, P. C. Knag and R. K. Krishnamurthy, "A 4096-Neuron 1M-Synapse 3.8PJ/SOP Spiking Neural Network with On-Chip STDP Learning and Sparse Weights in 10NM FinFET CMOS," IEEE Symposium on VLSI Circuits, 2018